\pdfoutput=1

\documentclass[11pt]{article}

\usepackage[]{EMNLP2023}

\usepackage{times}
\usepackage{latexsym}

\usepackage[T1]{fontenc}

\usepackage[utf8]{inputenc}

\usepackage{microtype}

\usepackage{inconsolata}
\usepackage{multirow}
\usepackage{graphicx}
\usepackage{float}

\usepackage{tabularx}
\newcolumntype{Y}{>{\arraybackslash}X}
\usepackage{ amssymb }
\usepackage{ wasysym }

\title{Single-Sentence Reader: A Novel Approach\\ for Addressing Answer Position Bias}

\author{Son Quoc Tran \\
  Computer Science \\
  Denison University, Granville, OH \\
  \texttt{tran\_s2@denison.edu} \\\And
  Matt Kretchmar \\
  Computer Science \\
  Denison University, Granville, OH \\
  \texttt{kretchmar@denison.edu} \\}

\begin{document}
\maketitle
\begin{abstract}
Machine Reading Comprehension (MRC) models tend to take advantage of spurious correlations (also known as dataset bias or annotation artifacts in the research community). Consequently, these models may perform the MRC task without fully comprehending the given context and question, which is undesirable since it may result in low robustness against distribution shift. The main focus of this paper is answer-position bias, where a significant percentage of training questions have answers located solely in the first sentence of the context. We propose a Single-Sentence Reader as a new approach for addressing answer position bias in MRC. Remarkably, in our experiments with six different models, our proposed Single-Sentence Readers trained on biased dataset achieve results that nearly match those of models trained on normal dataset, proving their effectiveness in addressing the answer position bias. Our study also discusses several challenges our Single-Sentence Readers encounter and proposes a potential solution.

\footnote{Our code is publicly available at: \url{https://github.com/sonqt/single-sentence-reader}.}

\end{abstract}
\section{Introduction}
\begin{figure}[ht]
\centering
\includegraphics[width =7cm]{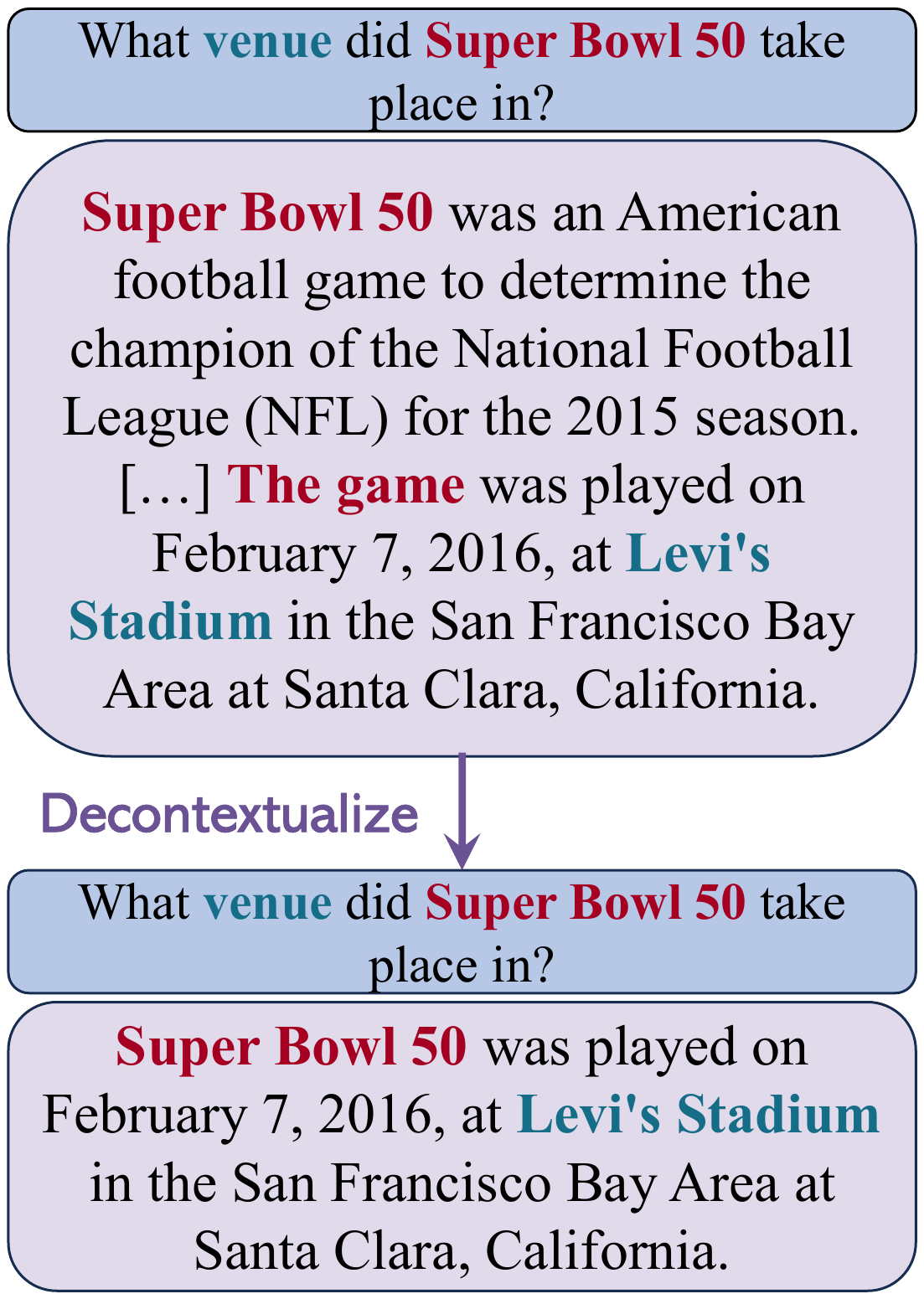}
\caption{An illustration of decontextualization in Machine Reading Comprehension. This process supports Single-Sentence Reader, our proposed approach, for addressing answer position bias in Machine Reading Comprehension.}
\label{fig:example}
\end{figure}

With the development of recent deep learning models, Machine Reading Comprehension (MRC) systems have achieved new state-of-the-art (SOTA) performances, matching or exceeding human-level standards on many benchmarks. However, recent works have indicated that multiple MRC benchmarks have annotation artifacts that may overestimate the assessment of comprehension ability of current SOTA MRC models \cite{mccoy-etal-2019-right, gardner-etal-2020-evaluating, sugawara-etal-2018-makes, gururangan-etal-2018-annotation}.

Machine Reading Comprehension (MRC) models tend to take advantage of spurious correlations (also known as dataset bias or annotation artifacts in the research community) \cite{shinoda2022shortcut}. These correlations in the training data lead to learning shortcuts of models during the training phase. Consequently, these models may perform the MRC task without fully comprehending the given context and question \cite{mccoy-etal-2019-right, gardner-etal-2020-evaluating}. This behavior of MRC models is undesirable since it may result in low robustness against distribution shift \cite{khashabi-etal-2020-bang, pmlr-v119-miller20a, sulem-etal-2021-know-dont} or adversarial attack \cite{jia-liang-2017-adversarial, bartolo-etal-2020-beat}.

The focus of this paper is answer-position bias, where a significant percentage of training questions have answers located in the first sentence of the context \cite{ko-etal-2020-look}. We observe that the main reason causing the answer-position bias is that when the model is trained with a biased train set, MRC models will always look to the first sentence to find the answer without considering other sentences in the given context. From this observation, we propose Single-Sentence Reader as an approach for addressing answer-position bias. Single-Sentence Reader first leverages the knowledge of models from decontextualization task \cite{choi-etal-2021-decontextualization} 
to rewrite a sentence to be interpretable out of its original context while preserving its meaning. Therefore, the problem of addressing answer position now reduces to the problem of maximizing the probability of the correct answer among predictions from each sentence.

However, later in this paper, we show that if the model is not trained to produce low probability scores for spans from sentences that do not contain an answer, comparing probability scores of predictions would yield undesirable performances. In order to train our models to recognize the unanswerability of a question given a sentence, we include a number of auto-created unanswerable questions in the training set. The unanswerable questions we use in this experiment are auto-created unanswerable questions, which are automatically collected using TF-IDF, a heuristic retriever, inspired by \citet{clark-gardner-2018-simple, chen-etal-2017-reading}.

In order to evaluate the effectiveness of our proposed technique, we systematically explore the performance differences of six pre-trained language models using both traditional and single-sentence approaches.

Our contributions in this paper are summarized as follows:
\begin{enumerate}
    \item We propose Single-Sentence Reader, an approach for addressing answer-position bias. Given only the biased training data, the results of our experiments indicate significant improvements, with approximately 50 F1 points gain compared to traditional models when evaluated on the anti-biased test set. Furthermore, we highlight the crucial role of incorporating unanswerable questions in the training set for the success of the Single-Sentence Reader approach. 
    \item We also discuss three main challenges that Single-Sentence Readers encounter, including the poor quality of testing samples, failures of the decontextualizing model, and challenges arising from single-sentence settings. Additionally, we propose a potential solution to enhance the performance of the Single-Sentence Reader further.
\end{enumerate}

\section{Related Work}
\subsection{Machine Reading Comprehension Bias}
Shortcut learning by Deep Learning models \cite{Geirhos_2020} has received significant attention within the research community in recent years. This is primarily due to the adverse impact that shortcut learning has on the performance of neural models, especially when tested on out-of-domain test sets and against adversarial attacks. This issue also affects models designed for MRC. Although MRC models have achieved human-level performance on some benchmarks \cite{rajpurkar-etal-2016-squad}, they lack robustness to challenging test sets such as adversarial attacks \cite{jia-liang-2017-adversarial, tran-etal-2023-impacts}, adversarially annotated questions \cite{bartolo-etal-2020-beat}, answers in unseen positions \cite{ko-etal-2020-look}, and natural perturbations \cite{gardner-etal-2020-evaluating, khashabi-etal-2020-bang}.

There are multiple attempts by the research community to further understand the learning shortcuts of MRC models. In a study by \citet{lai-etal-2021-machine}, they revealed that MRC models tend to learn from shortcut questions ealier than from challenging ones. Another work conducted by \citet{shinoda2022shortcut} involved behavioral tests with biased training sets focusing on the learnability of shortcuts. This study highlighted that the degree of learnability of a type of shortcut impacts the proportion of anti-shortcut examples necessary to achieve comparable performance.
\subsection{Unanswerable Questions in MRC}
In the direction of unanswerable question in MRC research, early efforts by \citet{levy-etal-2017-zero} involved redefining the BiDAF model \cite{DBLP:journals/corr/SeoKFH16}, enabling the model to determine whether a provided question is unanswerable. TThrough this redefinition, \citet{DBLP:journals/corr/SeoKFH16} effectively utilize capabilities of Machine Reading Comprehension (MRC) models to extract relationships in zero-shot settings.

Subsequently, \citet{rajpurkar-etal-2018-know} introduced a crowdsourcing methodology for annotating unanswerable questions. This initiative led to the creation of the SQuAD 2.0 dataset, designed for Extractive Question Answering. This milestone served as inspiration for analogous endeavors in other languages, such as French \cite{https://doi.org/10.48550/arxiv.2109.13209} and Vietnamese \cite{https://doi.org/10.48550/arxiv.2203.11400}. 
Recently, \citet{tran-etal-2023-impacts} demonstrate the usefulness of adversarial unanswerable questions in improving the robustness of models against adversarial attacks. However, \citet{sulem-etal-2021-know-dont} highlight a significant limitation of SQuAD 2.0: models trained on this dataset tend to exhibit subpar performance when tested on samples from domains beyond their training data. 

In addition to the adversarially-crafted unanswerable questions proposed by \citet{rajpurkar-etal-2018-know}, Natural Question \cite{kwiatkowski-etal-2019-natural} and Tydi QA \cite{clark-etal-2020-tydi} propose more naturally constructed unanswerable questions. While recent language models surpass human performances on adversarial unanswerable questions of SQuAD 2.0, natural unanswerable questions in Natural Question and Tidy QA remain challenging \cite{asai-choi-2021-challenges}.

\begin{figure*}[ht]
\centering
\includegraphics[width =\textwidth]{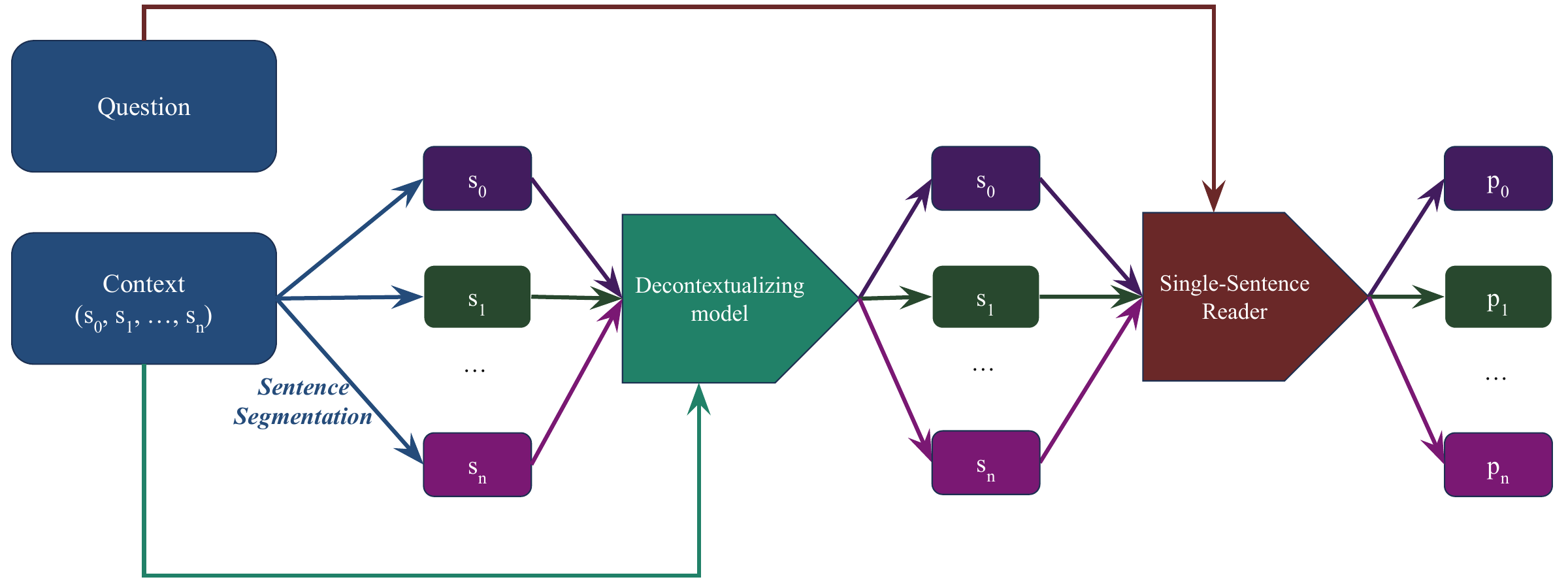}
\caption{Illustration of the proposed Single-Sentence Reader. The sentences in a given context are denoted as $s_0$, $s_1$, ..., $s_n$, and the corresponding predictions by Single-Sentence Reader are $p_0$, $p_1$, ..., $p_n$. The final precision is the prediction $p_k$ such that the confidence (probability outputted by softmax function) of this prediction is highest among $n$ predictions.}
\label{fig:pipeline}
\end{figure*}
\section{Experimental Setup}
\subsection{Dataset and Model}

Our experiment uses SQuAD 1.1 (Stanford Question Answering Dataset) \cite{rajpurkar-etal-2016-squad}, a benchmark dataset for MRC tasks. It was released in 2016 and revised in 2018 \cite{rajpurkar-etal-2018-know}. The SQuAD 1.1 dataset comprises around $100,000$ question-answer pairs on $536$ Wikipedia articles. The questions are written by humans and are based on the corresponding paragraphs in the articles. The task for MRC models of the dataset is to read a given paragraph (context) and extract a span from the given context to answer the corresponding questions accurately. The metrics used to evaluate the MRC models on SQuAD are 
\begin{itemize}
    \item \textbf{EM} (Exact Match) measures the percentage of predictions by models that exactly match the ground truth provided by the dataset.
    \item \textbf{F1-score} measures the average overlap between predictions and gold answers.
\end{itemize}
Please refer to \cite{rajpurkar-etal-2016-squad} for more details.

In this paper, we experiment with three pre-trained state-of-the-art transformer-based \cite{10.5555/3295222.3295349} models BERT \cite{devlin-etal-2019-bert}, RoBERTa \cite{DBLP:journals/corr/abs-1907-11692}, and SpanBERT \cite{joshi-etal-2020-spanbert}) in our work. 
\textbf{BERT} is trained on English Wikipedia plus BookCorpus with the pre-training tasks of masked language modeling (MLM) and next sentence prediction (NSP). Later, \citet{DBLP:journals/corr/abs-1907-11692} revealed substantial under-training of BERT. As a response, \citet{DBLP:journals/corr/abs-1907-11692} then developed \textbf{RoBERTa} by enhancing BERT's capabilities through increasing pre-training time and dataset size. \textbf{SpanBERT} \cite{joshi-etal-2020-spanbert} aiming to enhance text representation of model, replaces next sentence prediction with a span boundary objective.

Each of these three models has two versions: base and large. Our study uses all six of these models.

\subsection{Training Set}
\label{sec:answer-position-train-set}
\begin{table}[ht]
\centering
\resizebox{7cm}{!}{%
\begin{tabular}{|l|l|}
\hline
\textbf{Training Set} & \textbf{Characteristics} \\ \hline
Biased & \begin{tabular}[c]{@{}l@{}} \textcolor{purple}{$\fullmoon$ Biased}\\ \textcolor{teal}{$\square$ Full Paragraph}\\ \textcolor{violet}{$\vartriangle$ \textbf{No} Unanswerable}\end{tabular} \\ \hline
\begin{tabular}[c]{@{}l@{}}Single-Sentence \\ (w/o unanswerable)\end{tabular} & \begin{tabular}[c]{@{}l@{}}\textcolor{purple}{$\fullmoon$ Biased}\\ \textcolor{teal}{ $\square$ First Sentence Only}\\ \textcolor{violet}{$\vartriangle$  \textbf{No} Unanswerable}\end{tabular} \\ \hline
\begin{tabular}[c]{@{}l@{}}Single-Sentence \\ (with unanswerable)\end{tabular} & \begin{tabular}[c]{@{}l@{}}\textcolor{purple}{$\fullmoon$ Biased}\\ \textcolor{teal}{$\square$ First Sentence Only}\\ \textcolor{violet}{$\vartriangle$  Unanswerable}\end{tabular} \\ \hline
Anti-Biased & \begin{tabular}[c]{@{}l@{}}\textcolor{purple}{$\fullmoon$ \textbf{Anti}-Biased}\\ \textcolor{teal}{$\square$ Full Paragraph}\\ \textcolor{violet}{$\vartriangle$ \textbf{No} Unanswerable}\end{tabular} \\ \hline
\end{tabular}
}

\caption{Key characteristics of four different training sets used in our experiments.}
\label{tab:summary-training}
\end{table}

Table \ref{tab:summary-training} summarizes the key characteristics of four different training sets we utilize in our experiments. The biased training set serves as the baseline for demonstrating the effectiveness of our proposed method. On the other hand, models fine-tuned using the training set under normal settings offer the desired performance we aim to achieve with our proposed method. In order to study the importance of unanswerable questions in our proposed method, we fine-tune Single-Sentence Readers using two different sets: one with unanswerable questions and the other without.
\subsubsection*{Biased Set}

In order to create the biased training set for our controlled experiment, we utilize SpaCy's pipeline \footnote{https://github.com/explosion/spaCy} to perform sentence segmentation on contexts of questions. Sentence segmentation, also known as sentence boundary detection, is the process of dividing a continuous stream of text into individual sentences. Next, we examine whether the answer annotated by the crowdsourced workers of SQuAD 1.1 is present in the first sentence of the context. If the answer locates in the first sentence, we include it in our Answer-Position \textbf{Biased} training set. The Biased training set includes $27,929$ questions (see Section \ref{sec:answer-position-survey}).
\subsubsection*{Single-Sentence without Unanswerable}
The training set for Single-Sentence Reader without unanswerable is then built from the Biased training set. The key difference is that contexts in this new training set are created by deleting all sentences after the first sentence.
\subsubsection*{Single-Sentence with Unanswerable}
This training set includes all answerable questions from the training set of Single-Sentence \textbf{without} Unanswerable. We augment this training set with automatically created unanswerable questions.

Inspired by the idea of \citet{clark-gardner-2018-simple}, we use a heuristic algorithm to match a question with a new context that differs from the original context paired to the given question. Specifically, we calculate TF-IDF (term frequency-inverse document frequency) scores of each question and each \textbf{first} sentence of context in SQuAD 1.1 \cite{rajpurkar-etal-2016-squad}. 

For each question, we retain the five highest-ranked sentences (not the sentences that contain the ground truth answer) in a large pool of unanswerable question pairs (question and first sentence); then, we randomly select $n$ questions from this pool. In this paper, following the ratio of SQuAD 2.0, we set the number of unanswerable questions to be half of the number of answerable questions in the training set. We then combine this set of unanswerable questions with answerable single-sentence training set to create the training set for Single-Sentence Reader with unanswerable.

During the training phase, we train the Single-Sentence Readers to output empty string for unanswerable questions.
\subsubsection*{Normal Settings}
In order to establish a desired baseline for our Single-Sentence Readers, we fine-tune models following the traditional process of EQA using the normal training set. This set comprises both biased and anti-biased samples. To ensure a fair comparison, the size of this set matches that of the biased set, totaling $27,929$ samples.
\subsection{Testing Set}
Given 4 training sets, we evaluate all our models on 2 testing sets.
\subsubsection*{Biased} 
This set comprises biased samples from the development set of SQuAD 1.1, and it is extracted through a similar process used for obtaining Biased training set from the development set. The Biased testing set includes $3,435$ questions.
\subsubsection*{Anti-Biased} 
This set comprises anti-biased samples from the development set of SQuAD 1.1, which means that the answers to these questions are not in the first sentence of the corresponding context. The Anti-Biased testing set includes $7,135$ questions.
\subsection{Answer-Position Bias in SQuAD}
\label{sec:answer-position-survey}
\begin{table}[ht]
\resizebox{\linewidth}{!}{%
\begin{tabular}{lccc}
\hline
 & Train & Development & Total \\ \hline
Biased & 27,929 & \multicolumn{1}{c}{3,435} & 31,364 \\
Anti-Biased & 59,669 & \multicolumn{1}{c}{7,135} & 66,804 \\ \hline
Total & 87,598 & \multicolumn{1}{c}{10,570} & 98,168 \\ \hline
\end{tabular}%
}
\caption{Investigation of Answer-Position Bias in SQuAD 1.1. Biased samples have answers in the first sentence of the context.}
\label{tab:squad-bias}
\end{table}
We survey the Answer-Position Bias in the SQuAD 1.1 dataset \cite{rajpurkar-etal-2016-squad} for building the Biased and Anti-Biased sets from training and development set of the original dataset. We follow the process outlined in Section \ref{sec:answer-position-train-set} and report the specific numbers of Biased and Anti-Biased questions in Table \ref{tab:squad-bias}. 
\subsection{Details for MRC Model Training}
This work uses the base and large versions for all considered pre-trained models. We train all MRC models using a batch size of 8 for 2 epochs. The maximum sequence length is set to 384 tokens. We use the AdamW optimizer \cite{loshchilov2018decoupled} with an initial learning rate of $2\cdot 10^{-5}$, and $\beta_1 = 0.9$, $\beta_2 = 0.999$. We fine-tuned all models on a single NVIDIA GeForce RTX 3080.
\section{Single-Sentence Reader}

\begin{table*}[!ht]
\centering
\resizebox{12cm}{!}{%
\begin{tabular}{lllcc|cc}
\hline
 &  & \multicolumn{1}{r}{\textbf{\textit{Test}} $\rightarrow$} & \multicolumn{2}{c|}{\textbf{Biased}} & \multicolumn{2}{c}{\textbf{Anti-Biased}} \\
 &  & \textit{\textbf{Train}} $\downarrow$ & EM & F1 & EM & F1 \\ \hline
\multirow{8}{*}{\textbf{BERT}} & \multirow{4}{*}{\textbf{base}} & Biased & \textbf{78.9} & \textbf{87.3} & 12.2 & 17.5 \\ \cline{3-3}
 &  & \begin{tabular}[c]{@{}l@{}}Single-Sentence\\ (w/o unanswerable)\end{tabular} & 55.8 & 61.5 & 41.2 & 46.6 \\ \cline{3-3}
 &  & \textbf{\begin{tabular}[c]{@{}l@{}}Single-Sentence\\ (with unanswerable)\end{tabular}} & 70.2 & 77.4 & \textbf{60.7} & \textbf{68.4} \\ \cline{3-3}
 &  & \textit{Normal Settings} & \textit{78.7} & \textit{86.3} & \textit{75.9} & \textit{84.8} \\ \cline{2-7} 
 & \multirow{4}{*}{\textbf{large}} & Biased & \textbf{83.8} & \textbf{90.9} & 12.4 & 17.5 \\ \cline{3-3}
 &  & \begin{tabular}[c]{@{}l@{}}Single-Sentence\\ (w/o unanswerable)\end{tabular} & 60.1 & 65.6 & 47.2 & 52.3 \\ \cline{3-3}
 &  & \textbf{\begin{tabular}[c]{@{}l@{}}Single-Sentence\\ (with unanswerable)\end{tabular}} & 77.0 & 83.1 & \textbf{65.5} & \textbf{72.8} \\ \cline{3-3}
 &  & \textit{Normal Settings} & \textit{82.3} & \textit{89.4} & \textit{81.0} & \textit{89.0} \\ \hline
\multirow{8}{*}{\textbf{RoBERTa}} & \multirow{4}{*}{\textbf{base}} & Biased & \textbf{85.1} & \textbf{91.6} & 28.3 & 34.3 \\ \cline{3-3}
 &  & \begin{tabular}[c]{@{}l@{}}Single-Sentence\\ (w/o unanswerable)\end{tabular} & 61.7 & 66.9 & 47.0 & 52.2 \\ \cline{3-3}
 &  & \textbf{\begin{tabular}[c]{@{}l@{}}Single-Sentence\\ (with unanswerable)\end{tabular}} & 77.4 & 83.5 & \textbf{67.7} & \textbf{74.3} \\ \cline{3-3}
 &  & \textit{Normal Settings} & \textit{84.2} & \textit{90.7} & \textit{83.2} & \textit{90.3} \\ \cline{2-7} 
 & \multirow{4}{*}{\textbf{large}} & Biased & \textbf{88.0} & \textbf{93.8} & 16.5 & 22.2 \\ \cline{3-3}
 &  & \begin{tabular}[c]{@{}l@{}}Single-Sentence\\ (w/o unanswerable)\end{tabular} & 64.6 & 69.7 & 50.3 & 55.4 \\ \cline{3-3}
 &  & \textbf{\begin{tabular}[c]{@{}l@{}}Single-Sentence\\ (with unanswerable)\end{tabular}} & 81.7 & 87.4 & \textbf{71.1} & \textbf{77.6} \\ \cline{3-3}
 &  & \textit{Normal Settings} & \textit{88.3} & \textit{93.5} & \textit{87.0} & \textit{93.2} \\ \hline
\multirow{8}{*}{\textbf{SpanBERT}} & \multirow{4}{*}{\textbf{base}} & Biased & \textbf{82.6} & \textbf{90.3} & 19.8 & 26.0 \\ \cline{3-3}
 &  & \begin{tabular}[c]{@{}l@{}}Single-Sentence\\ (w/o unanswerable)\end{tabular} & 55.0 & 60.8 & 40.9 & 46.3 \\ \cline{3-3}
 &  & \textbf{\begin{tabular}[c]{@{}l@{}}Single-Sentence\\ (with unanswerable)\end{tabular}} & 74.7 & 81.7 & \textbf{63.9} & \textbf{71.6} \\ \cline{3-3}
 &  & \textit{Normal Settings} & \textit{81.9} & \textit{89.2} & \textit{80.8} & \textit{89.0} \\ \cline{2-7} 
 & \multirow{4}{*}{\textbf{large}} & Biased & \textbf{86.2} & \textbf{92.6} & 14.4 & 20.2 \\ \cline{3-3}
 &  & \begin{tabular}[c]{@{}l@{}}Single-Sentence\\ (w/o unanswerable)\end{tabular} & 61.0 & 66.5 & 46.7 & 52.0 \\ \cline{3-3}
 &  & \textbf{\begin{tabular}[c]{@{}l@{}}Single-Sentence\\ (with unanswerable)\end{tabular}} & 79.7 & 85.9 & \textbf{68.8} & \textbf{75.7} \\ \cline{3-3}
 &  & \textit{Normal Settings} & \textit{86.5} & \textit{92.5} & \textit{85.9} & \textit{92.5} \\ \hline
\multicolumn{2}{l}{\multirow{4}{*}{\textbf{Average}}} & Biased & \textbf{84.1}$_{\pm 3.2}$ & \textbf{91.1}$_{\pm 5}$ & 17.3$_{\pm 6.1}$ & 22.9$_{\pm 6.4}$ \\ \cline{3-3}
\multicolumn{2}{l}{} & \begin{tabular}[c]{@{}l@{}}Single-Sentence\\ (w/o unanswerable)\end{tabular} & 59.7$_{\pm 3.7}$ & 65.2$_{\pm 3.4}$ & 45.6$_{\pm 3.7}$ & 50.8$_{\pm 3.6}$ \\ \cline{3-3}
\multicolumn{2}{l}{} & \textbf{\begin{tabular}[c]{@{}l@{}}Single-Sentence\\ (with unanswerable)\end{tabular}} & 76.8$_{\pm 4.0}$ & 81.2$_{\pm 3.5}$ & \textbf{66.3}$_{\pm 3.7}$ & \textbf{73.4}$_{\pm 3.2}$ \\ \cline{3-3}
\multicolumn{2}{l}{} & \textit{Normal Settings} & \textit{83.7}$_{\pm 3.4}$ & \textit{90.3}$_{\pm 2.6}$ & \textit{82.3}$_{\pm 4}$ & \textit{89.8}$_{\pm 3.0}$ \\ \hline
\end{tabular}
}
\caption{The average of performance (EM and F1 scores) by 6 models fine-tuned using different training sets. We also report the average performance with standard deviation of these 6 models. While \textit{Biased} and \textit{Normal Settings} models follow the traditional process of MRC models during the testing phase, Single-Sentence Readers fine-tuned follow the algorithm outlined in \textsection \ref{sec:method}.}
\label{tab:final_result}
\end{table*}

\label{sec:method}
This section describes the inference pipeline for Single-Sentence Readers (both with and without unanswerable).

Figure \ref{fig:pipeline} illustrates the workings of the pipeline in the testing phase. Given a Single-Sentence Readers, we perform the following step to get the predicted answer from the model:
\begin{enumerate}
    \item Segment the given context (corresponding to the question) into sentences ($s_1, s_2, ..., s_n$). Then, we decontextualize sentences to enable each to stand alone as a new context (see Figure \ref{fig:example}). In our experiments, we use the model T5-base \cite{JMLR:v21:20-074} fine-tuned on the Decontextualization task \cite{choi-etal-2021-decontextualization}.
    \item Independently get a prediction $p_k$ from the model on each sentence $s_k$. We then have predictions $(p_1, p_2,...p_n)$ corresponding to ($s_1, s_2, ..., s_n$).
    \item The final answer for the given question is the non-empty prediction in $(p_1, p_2,...p_n)$ with the highest probability (output of softmax function).
    \item A possible problem with the Single-Sentence Readers fine-tuned \textbf{with} unanswerable questions is that they may give empty predictions for every sentence that is taken out of its original context. If there is no non-empty prediction from Single-Sentence Readers fine-tuned \textbf{with} unanswerable, the final prediction of this model will then be an empty string.
\end{enumerate}

\section{Results}

\label{sec:debias-w-unanswerable}
Table \ref{tab:final_result} presents the performance of 6 models fine-tuned on 4 different datasets. Firstly, we observe that \textbf{biased} models fine-tuned using a training set with only biased questions show poor performance on anti-biased samples ($17.3$ EM and $22.9$ F1 on average). This observation highlights the need for a comprehensive study on methods for addressing answer-position bias. 

We then investigate the performance of \textbf{Single-Sentence Readers} fine-tuned \textbf{without unanswerable} questions. The results show that these models' performance significantly improves on anti-biased samples compared to the biased baselines. These improvements show the potential of the Single-Sentence approach in addressing answer-position bias in EQA. However, the performance of these models shows a significant decline (about $25$ EM and F1) compared to the biased baselines on biased samples. 

Our results show that the \textbf{Single-Sentence Readers} fine-tuned \textbf{without unanswerable} questions perform poorly because they are not trained to assign low probability scores to sentences that do not contain an answer. This makes the comparison between probability scores of predictions in different sentences unreliable. To address this issue, we train \textbf{Single-Sentence Readers with unanswerable questions} that are automatically created. This significantly improves the performance of Single-Sentence Readers on both biased (F1 score increased from $65.2$ to $81.2$) and anti-biased samples (F1 score increased from $50.8$ to $73.4$).

However, when comparing the performance of Single-Sentence Readers with the traditional EQA models fine-tuned under normal settings, we observe a considerable gap between the performance of Single-Sentence Readers and that of models under normal settings. This gap is headroom for further improvements. In the next section, we investigate the challenges that Single-Sentence Readers encounter.
\section{Discussion}
\begin{table*}[ht]
\centering
\resizebox{14cm}{!}{%
\begin{tabular}{|l|l|}
\hline
\multicolumn{1}{|c|}{\textbf{Example}} & \multicolumn{1}{c|}{\textbf{Challenge}} \\ \hline
\begin{tabular}[c]{@{}l@{}}\textbf{Question}: How much did it cost to build the stadium where Super Bowl 50\\  was played?\\ \textbf{Context}: On May 21, 2013, NFL owners at their spring meetings in Boston\\ voted and awarded the game to Levi's Stadium. The \textcolor{red}{\textbf{\$1.2 billion}} stadium\\ opened in 2014. [...]\end{tabular} & \textbf{\begin{tabular}[c]{@{}l@{}}Missing information\\ from the original context\end{tabular}} \\ \hline
\begin{tabular}[c]{@{}l@{}}\textbf{Question}: Which team won Super Bowl 50?\\ \textbf{Context}: Super Bowl 50 was an American football game to determine the\\ champion of the National Football League (NFL) for the 2015 season. The\\ American Football Conference (AFC) champion \textcolor{red}{\textbf{Denver Broncos}} defeated \\ the National Football Conference (NFC) champion Carolina Panthers 24-10\\ to earn their third Super Bowl title. [...]\end{tabular} & \textbf{\begin{tabular}[c]{@{}l@{}}Missing information\\ due to Single-Sentence \\ settings\end{tabular}} \\ \hline
\begin{tabular}[c]{@{}l@{}}\textbf{Question}: When did Levi's stadium open to the public?\\ \textbf{Context}: On May 21, 2013, NFL owners at their spring meetings in Boston\\ voted and awarded the game to Levi's Stadium. The \$1.2 billion stadium \\ opened in \textcolor{red}{\textbf{2014}}. [...]\end{tabular} & \textbf{\begin{tabular}[c]{@{}l@{}}Missing information\\ due to Single-Sentence \\ settings\end{tabular}} \\ \hline
\begin{tabular}[c]{@{}l@{}}\textbf{Question}: How many appearances have the Denver Broncos made in the Super\\ Bowl?\\ \textbf{Context}: The Panthers [...]. The Broncos [...] denied the New England Patriots\\ a chance to defend their title from Super Bowl XLIX by defeating them 20-8 in\\ the AFC Championship Game. \textcolor{teal}{\textit{They joined the Patriots, Dallas Cowboys, and}}\\ \textcolor{teal}{\textit{Pittsburgh Steelers as one of four teams that have made \textbf{eight} appearances in}}\\ \textcolor{teal}{\textit{the Super Bowl.}}\\ \textbf{Decontextualization}:\textcolor{red}{The Carolina Panthers} \textcolor{teal}{\textit{joined the Patriots , Dallas}}\\ \textcolor{teal}{\textit{Cowboys , and Pittsburgh Steelers as one of four teams that have made eight}}\\ \textcolor{teal}{\textit{appearances in the Super Bowl.}}\end{tabular} & \textbf{\begin{tabular}[c]{@{}l@{}}Failures of \\ Decontextualizing \\ Model\end{tabular}} \\ \hline
\begin{tabular}[c]{@{}l@{}}\textbf{Question}: How many sacks did Derek Wolfe register?\\ \textbf{Context}: The Broncos' defense ranked first in the NFL yards allowed (4,530)\\ for the first time in franchise history, and fourth in points allowed (296). \\ \textcolor{teal}{\textit{Defensive ends Derek Wolfe and Malik Jackson each had \textbf{5}$\frac{1}{2}$ sacks.}}[...]\\ \textbf{Decontextulization}: \textcolor{red}{The Broncos' defense ranked first in the NFL yards} \\ \textcolor{red}{allowed (4,530) for the first time in franchise history, and fourth in points}\\ \textcolor{red}{allowed (296)}.\end{tabular} & \textbf{\begin{tabular}[c]{@{}l@{}}Failures of \\ Decontextualizing \\ Model\end{tabular}} \\ \hline
\end{tabular}
}
\caption{Examples of challenges that Single-Sentence Reader encounters. The \textbf{ground truth answers} are highlighted in \textbf{bold} in the corresponding context. The \textcolor{teal}{\textit{decontextualizing target sentences}} are highlighted in \textcolor{teal}{\textit{teal}} color.}
\label{tab:challenge-example}
\end{table*}

In addition to the challenges the original SQuAD 1.1 dataset poses, our Single-Sentence Reader also encounters new challenges due to multiple factors. In this Section, we first discuss the challenges we observe through our comprehensive study and then propose a potential solution to improve the performance of Single-Sentence Readers further.
\subsection{Challenges}
\subsubsection*{Missing information from original context}
This challenge arises because of the settings of the anntation process of SQuAD 1.1. This challenge is illustrated in the first example in Table \ref{tab:challenge-example}. As crowd-workers are presented with paragraphs in a Wikipedia page in order, they can infer that ``the game'' in the presented context refers to ``Super Bowl 50''. However, in the presented context, ``Super Bowl 50'' is not mentioned. Therefore, this causes a significant challenge for Single-Sentence Reader, as this type of model has the ability to recognize the unanswerability of the given questions.
\subsubsection*{Missing information due to Single-Sentence settings}
Although every sentence used in our Single-Sentence settings is decontextualized before being used as the context for Single-Sentence Reader, there are still cases where the target sentence does not have enough key information to answer the corresponding question. For example, in the second example of Table \ref{tab:challenge-example}, we know that Denver Broncos defeated Carolina Panthers and won the match. However, if we take this sentence out of the original context, there is no clue indicating whether the referred match in this sentence is Super Bowl 50. 
\subsubsection*{Failures of Decontextualizing Model}
The decontextualizing model may encounter difficulties in accurately decontextualizing sentences within a paragraph, especially when the paragraph is excessively complex. This challenge is illustrated in the fourth example in Table \ref{tab:challenge-example}. As Carolina Panthers and Denver Broncos both appear in the context, the decontextualizing model produces an incorrect coreference resolution (\textbf{they-Panthers} instead of \textbf{they-Broncos}).

There are also cases such that we lose the correct answer for a corresponding question after decontextualization. This problem is illustrated in the last example in Table \ref{tab:challenge-example}. The decontextualizing model mistakenly copies the whole first sentence in an effort to decontextualize the second sentence. Therefore, after decontextualizing the second sentence, we lose the ground-truth answer to the question ``How many sacks did Derek Wolfe register?''.

\subsection{Potential Solution}
We propose a solution for the challenges of missing information from the original context and missing information due to Single-Sentence settings. Firstly, as discussed in the previous Section, these challenges arise partly due to the task of unanswerability recognition of Single-Sentence Readers. Therefore, comparing Single-Sentence Readers with models and normal settings is unfair because models under normal settings cannot recognize an unanswerable question. From this observation, we apply the technique force-to-answer \cite{tran-etal-2023-impacts} to further boost the performance of the Single-Sentence Reader. 
\subsubsection{Single-Sentence Reader with Force-To-Answer}
We reformulate the inferencing pipeline of Single-Sentence Reader to ensure we can obtain a non-empty answer for every question.
\begin{enumerate}
    \item Segment the given context (corresponding to the question) into sentences ($s_1, s_2, ..., s_n$). Decontextualize sentences to enable each of them to stand alone as a new context. In our experiments, we use the model T5-base fine-tuned on the Decontextualization task \cite{choi-etal-2021-decontextualization}.
    \item Independently get \textbf{2 highest predictions} ($p^1_k$ and $p^2_k$) from the model on each sentence $s_k$. We then have predictions $(p_1^1, p_1^2, p_2^1, p_2^2,..., p_n^1, p_n^2)$ corresponding to ($s_1, s_2, ..., s_n$).
    \item Then, the final answer for the given question is the non-empty prediction in $(p_1^1, p_1^2, p_2^1, p_2^2,..., p_n^1, p_n^2)$ with the highest probability (output of softmax function).
\end{enumerate}
\subsubsection{Result}
\begin{table}[ht]
\resizebox{\linewidth}{!}{%
\begin{tabular}{cc|cc|cc}
\hline
 &  & \multicolumn{2}{c|}{\textbf{Biased}} & \multicolumn{2}{c}{\textbf{Anti-Biased}} \\
 &  & EM & F1 & EM & F1 \\ \hline
\multirow{2}{*}{\textbf{BERT}} & base & 73 & 80.8 & 63.8 & 72.5 \\
 & large & 79.2 & 85.6 & 68.4 & 76.4 \\ \hline
\multirow{2}{*}{\textbf{RoBERTa}} & base & 79.6 & 86 & 71.1 & 78.5 \\
 & large & 83.3 & 89.2 & 73.7 & 80.8 \\ \hline
\multirow{2}{*}{\textbf{SpanBERT}} & base & 75.9 & 83.2 & 66.2 & 74.6 \\
 & large & 80.8 & 87.3 & 70.7 & 78.2 \\ \hline
\multicolumn{2}{c|}{\textbf{Average}} & 78.6$_{\pm 3.3}$ & 85.4$_{\pm 2.7}$ & 69$_{\pm 3.3}$ & 76.8$_{\pm 2.7}$ \\ \hline
\end{tabular}
}
\caption{Performance of Single-Sentence Readers with force-to-answer technique on Biased and Anti-Biased Answer Position Bias Sets. We also report the average performance with standard deviation of these six Readers.}
\label{tab:force2ans}
\end{table}
In Table \ref{tab:force2ans}, we report the performance of Single-Sentence Readers using the force-to-answer technique. From this experiment, we can observe that the performance of Single-Sentence Readers shows considerable improvement with the force-to-answer technique (85.4 F1 $-$ 81.2 F1 on \textbf{Biased} and 76.8 F1 $-$ 73.4 F1 on \textbf{Anti-Biased}).

Although the force-to-answer technique can help Single-Sentence Readers mitigate the challenges they encounter, applying this technique to our readers means removing the ability of unanswerability recognition. Therefore, this is a trade-off scenario to consider carefully, especially in real-world applications. 
\section{Conclusion}
In this work, we examine the answer-position bias in dataset SQuAD and introduce Single-Sentence Reader, a novel approach to address this bias. Our experiments demonstrate that the performance of Single-Sentence Readers nearly matches those of models fine-tuned on normal training sets. 

Besides, we also discuss 3 challenges that our Single-Sentence Readers encounter, including missing information from the original context, missing information due to Single-Sentence settings, and failures of the decontextualizing model. We propose force-to-answer as a potential solution for the first and second challenges to partly mitigate these challenges and further improve the performance of Single-Sentence Readers. However, applying force-to-answer on Single-Sentence Readers also means removing the ability of unanswerability recognition, which is crucial for our systems in real-world applications.

\section*{Limitations}

We recognize that contributions of this paper come with certain limitations. Firstly, in our conducted experiments throughout this paper, we predominantly leverage transformers-based pre-trained large language models. As a result, the applications of Single-Sentence Readers might not yield similar levels of improvements when applied to other models with different architectures such as Recurrent Neural Networks or Convolutional Neural Networks.

Secondly, the proposed Single-Sentence Readers relies on the availability of decontextualizing models. However, it is important to note that benchmarks for the decontextualization task are currently only accessible in the English language, following the work by \citet{choi-etal-2021-decontextualization}.

Lastly, we acknowledge that the proposed Single-Sentence Readers display a notable inefficiency during the inference phase when compared with traditional MRC models. This efficiency gap becomes evident in comparative evaluations of their computational performance.
\bibliography{anthology,custom}

\begin{thebibliography}{32}
\expandafter\ifx\csname natexlab\endcsname\relax\def\natexlab#1{#1}\fi

\bibitem[{Asai and Choi(2021)}]{asai-choi-2021-challenges}
Akari Asai and Eunsol Choi. 2021.
\newblock \href {https://doi.org/10.18653/v1/2021.acl-long.118} {Challenges in
  information-seeking {QA}: Unanswerable questions and paragraph retrieval}.
\newblock In \emph{Proceedings of the 59th Annual Meeting of the Association
  for Computational Linguistics and the 11th International Joint Conference on
  Natural Language Processing (Volume 1: Long Papers)}, pages 1492--1504,
  Online. Association for Computational Linguistics.

\bibitem[{Bartolo et~al.(2020)Bartolo, Roberts, Welbl, Riedel, and
  Stenetorp}]{bartolo-etal-2020-beat}
Max Bartolo, Alastair Roberts, Johannes Welbl, Sebastian Riedel, and Pontus
  Stenetorp. 2020.
\newblock \href {https://doi.org/10.1162/tacl_a_00338} {Beat the {AI}:
  Investigating adversarial human annotation for reading comprehension}.
\newblock \emph{Transactions of the Association for Computational Linguistics},
  8:662--678.

\bibitem[{Chen et~al.(2017)Chen, Fisch, Weston, and
  Bordes}]{chen-etal-2017-reading}
Danqi Chen, Adam Fisch, Jason Weston, and Antoine Bordes. 2017.
\newblock \href {https://doi.org/10.18653/v1/P17-1171} {Reading {W}ikipedia to
  answer open-domain questions}.
\newblock In \emph{Proceedings of the 55th Annual Meeting of the Association
  for Computational Linguistics (Volume 1: Long Papers)}, pages 1870--1879,
  Vancouver, Canada. Association for Computational Linguistics.

\bibitem[{Choi et~al.(2021)Choi, Palomaki, Lamm, Kwiatkowski, Das, and
  Collins}]{choi-etal-2021-decontextualization}
Eunsol Choi, Jennimaria Palomaki, Matthew Lamm, Tom Kwiatkowski, Dipanjan Das,
  and Michael Collins. 2021.
\newblock \href {https://doi.org/10.1162/tacl_a_00377} {Decontextualization:
  Making sentences stand-alone}.
\newblock \emph{Transactions of the Association for Computational Linguistics},
  9:447--461.

\bibitem[{Clark and Gardner(2018)}]{clark-gardner-2018-simple}
Christopher Clark and Matt Gardner. 2018.
\newblock \href {https://doi.org/10.18653/v1/P18-1078} {Simple and effective
  multi-paragraph reading comprehension}.
\newblock In \emph{Proceedings of the 56th Annual Meeting of the Association
  for Computational Linguistics (Volume 1: Long Papers)}, pages 845--855,
  Melbourne, Australia. Association for Computational Linguistics.

\bibitem[{Clark et~al.(2020)Clark, Choi, Collins, Garrette, Kwiatkowski,
  Nikolaev, and Palomaki}]{clark-etal-2020-tydi}
Jonathan~H. Clark, Eunsol Choi, Michael Collins, Dan Garrette, Tom Kwiatkowski,
  Vitaly Nikolaev, and Jennimaria Palomaki. 2020.
\newblock \href {https://doi.org/10.1162/tacl_a_00317} {{T}y{D}i {QA}: A
  benchmark for information-seeking question answering in typologically diverse
  languages}.
\newblock \emph{Transactions of the Association for Computational Linguistics},
  8:454--470.

\bibitem[{Devlin et~al.(2019)Devlin, Chang, Lee, and
  Toutanova}]{devlin-etal-2019-bert}
Jacob Devlin, Ming-Wei Chang, Kenton Lee, and Kristina Toutanova. 2019.
\newblock \href {https://doi.org/10.18653/v1/N19-1423} {{BERT}: Pre-training of
  deep bidirectional transformers for language understanding}.
\newblock In \emph{Proceedings of the 2019 Conference of the North {A}merican
  Chapter of the Association for Computational Linguistics: Human Language
  Technologies, Volume 1 (Long and Short Papers)}, pages 4171--4186,
  Minneapolis, Minnesota. Association for Computational Linguistics.

\bibitem[{Gardner et~al.(2020)Gardner, Artzi, Basmov, Berant, Bogin, Chen,
  Dasigi, Dua, Elazar, Gottumukkala, Gupta, Hajishirzi, Ilharco, Khashabi, Lin,
  Liu, Liu, Mulcaire, Ning, Singh, Smith, Subramanian, Tsarfaty, Wallace,
  Zhang, and Zhou}]{gardner-etal-2020-evaluating}
Matt Gardner, Yoav Artzi, Victoria Basmov, Jonathan Berant, Ben Bogin, Sihao
  Chen, Pradeep Dasigi, Dheeru Dua, Yanai Elazar, Ananth Gottumukkala, Nitish
  Gupta, Hannaneh Hajishirzi, Gabriel Ilharco, Daniel Khashabi, Kevin Lin,
  Jiangming Liu, Nelson~F. Liu, Phoebe Mulcaire, Qiang Ning, Sameer Singh,
  Noah~A. Smith, Sanjay Subramanian, Reut Tsarfaty, Eric Wallace, Ally Zhang,
  and Ben Zhou. 2020.
\newblock \href {https://doi.org/10.18653/v1/2020.findings-emnlp.117}
  {Evaluating models{'} local decision boundaries via contrast sets}.
\newblock In \emph{Findings of the Association for Computational Linguistics:
  EMNLP 2020}, pages 1307--1323, Online. Association for Computational
  Linguistics.

\bibitem[{Geirhos et~al.(2020)Geirhos, Jacobsen, Michaelis, Zemel, Brendel,
  Bethge, and Wichmann}]{Geirhos_2020}
Robert Geirhos, Jörn-Henrik Jacobsen, Claudio Michaelis, Richard Zemel,
  Wieland Brendel, Matthias Bethge, and Felix~A. Wichmann. 2020.
\newblock \href {https://doi.org/10.1038/s42256-020-00257-z} {Shortcut learning
  in deep neural networks}.
\newblock \emph{Nature Machine Intelligence}, 2(11):665--673.

\bibitem[{Gururangan et~al.(2018)Gururangan, Swayamdipta, Levy, Schwartz,
  Bowman, and Smith}]{gururangan-etal-2018-annotation}
Suchin Gururangan, Swabha Swayamdipta, Omer Levy, Roy Schwartz, Samuel Bowman,
  and Noah~A. Smith. 2018.
\newblock \href {https://doi.org/10.18653/v1/N18-2017} {Annotation artifacts in
  natural language inference data}.
\newblock In \emph{Proceedings of the 2018 Conference of the North {A}merican
  Chapter of the Association for Computational Linguistics: Human Language
  Technologies, Volume 2 (Short Papers)}, pages 107--112, New Orleans,
  Louisiana. Association for Computational Linguistics.

\bibitem[{Heinrich et~al.(2021)Heinrich, Viaud, and
  Belblidia}]{https://doi.org/10.48550/arxiv.2109.13209}
Quentin Heinrich, Gautier Viaud, and Wacim Belblidia. 2021.
\newblock \href {https://doi.org/10.48550/ARXIV.2109.13209} {Fquad2.0: French
  question answering and knowing that you know nothing}.

\bibitem[{Jia and Liang(2017)}]{jia-liang-2017-adversarial}
Robin Jia and Percy Liang. 2017.
\newblock \href {https://doi.org/10.18653/v1/D17-1215} {Adversarial examples
  for evaluating reading comprehension systems}.
\newblock In \emph{Proceedings of the 2017 Conference on Empirical Methods in
  Natural Language Processing}, pages 2021--2031, Copenhagen, Denmark.
  Association for Computational Linguistics.

\bibitem[{Joshi et~al.(2020)Joshi, Chen, Liu, Weld, Zettlemoyer, and
  Levy}]{joshi-etal-2020-spanbert}
Mandar Joshi, Danqi Chen, Yinhan Liu, Daniel~S. Weld, Luke Zettlemoyer, and
  Omer Levy. 2020.
\newblock \href {https://doi.org/10.1162/tacl_a_00300} {{S}pan{BERT}: Improving
  pre-training by representing and predicting spans}.
\newblock \emph{Transactions of the Association for Computational Linguistics},
  8:64--77.

\bibitem[{Khashabi et~al.(2020)Khashabi, Khot, and
  Sabharwal}]{khashabi-etal-2020-bang}
Daniel Khashabi, Tushar Khot, and Ashish Sabharwal. 2020.
\newblock \href {https://doi.org/10.18653/v1/2020.emnlp-main.12} {More bang for
  your buck: Natural perturbation for robust question answering}.
\newblock In \emph{Proceedings of the 2020 Conference on Empirical Methods in
  Natural Language Processing (EMNLP)}, pages 163--170, Online. Association for
  Computational Linguistics.

\bibitem[{Ko et~al.(2020)Ko, Lee, Kim, Kim, and Kang}]{ko-etal-2020-look}
Miyoung Ko, Jinhyuk Lee, Hyunjae Kim, Gangwoo Kim, and Jaewoo Kang. 2020.
\newblock \href {https://doi.org/10.18653/v1/2020.emnlp-main.84} {Look at the
  first sentence: Position bias in question answering}.
\newblock In \emph{Proceedings of the 2020 Conference on Empirical Methods in
  Natural Language Processing (EMNLP)}, pages 1109--1121, Online. Association
  for Computational Linguistics.

\bibitem[{Kwiatkowski et~al.(2019)Kwiatkowski, Palomaki, Redfield, Collins,
  Parikh, Alberti, Epstein, Polosukhin, Devlin, Lee, Toutanova, Jones, Kelcey,
  Chang, Dai, Uszkoreit, Le, and Petrov}]{kwiatkowski-etal-2019-natural}
Tom Kwiatkowski, Jennimaria Palomaki, Olivia Redfield, Michael Collins, Ankur
  Parikh, Chris Alberti, Danielle Epstein, Illia Polosukhin, Jacob Devlin,
  Kenton Lee, Kristina Toutanova, Llion Jones, Matthew Kelcey, Ming-Wei Chang,
  Andrew~M. Dai, Jakob Uszkoreit, Quoc Le, and Slav Petrov. 2019.
\newblock \href {https://doi.org/10.1162/tacl_a_00276} {Natural questions: A
  benchmark for question answering research}.
\newblock \emph{Transactions of the Association for Computational Linguistics},
  7:452--466.

\bibitem[{Lai et~al.(2021)Lai, Zhang, Feng, Huang, and
  Zhao}]{lai-etal-2021-machine}
Yuxuan Lai, Chen Zhang, Yansong Feng, Quzhe Huang, and Dongyan Zhao. 2021.
\newblock \href {https://doi.org/10.18653/v1/2021.findings-acl.85} {Why machine
  reading comprehension models learn shortcuts?}
\newblock In \emph{Findings of the Association for Computational Linguistics:
  ACL-IJCNLP 2021}, pages 989--1002, Online. Association for Computational
  Linguistics.

\bibitem[{Levy et~al.(2017)Levy, Seo, Choi, and
  Zettlemoyer}]{levy-etal-2017-zero}
Omer Levy, Minjoon Seo, Eunsol Choi, and Luke Zettlemoyer. 2017.
\newblock \href {https://doi.org/10.18653/v1/K17-1034} {Zero-shot relation
  extraction via reading comprehension}.
\newblock In \emph{Proceedings of the 21st Conference on Computational Natural
  Language Learning ({C}o{NLL} 2017)}, pages 333--342, Vancouver, Canada.
  Association for Computational Linguistics.

\bibitem[{Liu et~al.(2019)Liu, Ott, Goyal, Du, Joshi, Chen, Levy, Lewis,
  Zettlemoyer, and Stoyanov}]{DBLP:journals/corr/abs-1907-11692}
Yinhan Liu, Myle Ott, Naman Goyal, Jingfei Du, Mandar Joshi, Danqi Chen, Omer
  Levy, Mike Lewis, Luke Zettlemoyer, and Veselin Stoyanov. 2019.
\newblock \href {http://arxiv.org/abs/1907.11692} {Roberta: {A} robustly
  optimized {BERT} pretraining approach}.
\newblock \emph{CoRR}, abs/1907.11692.

\bibitem[{Loshchilov and Hutter(2019)}]{loshchilov2018decoupled}
Ilya Loshchilov and Frank Hutter. 2019.
\newblock \href {https://openreview.net/forum?id=Bkg6RiCqY7} {Decoupled weight
  decay regularization}.
\newblock In \emph{International Conference on Learning Representations}.

\bibitem[{McCoy et~al.(2019)McCoy, Pavlick, and Linzen}]{mccoy-etal-2019-right}
Tom McCoy, Ellie Pavlick, and Tal Linzen. 2019.
\newblock \href {https://doi.org/10.18653/v1/P19-1334} {Right for the wrong
  reasons: Diagnosing syntactic heuristics in natural language inference}.
\newblock In \emph{Proceedings of the 57th Annual Meeting of the Association
  for Computational Linguistics}, pages 3428--3448, Florence, Italy.
  Association for Computational Linguistics.

\bibitem[{Miller et~al.(2020)Miller, Krauth, Recht, and
  Schmidt}]{pmlr-v119-miller20a}
John Miller, Karl Krauth, Benjamin Recht, and Ludwig Schmidt. 2020.
\newblock \href {https://proceedings.mlr.press/v119/miller20a.html} {The effect
  of natural distribution shift on question answering models}.
\newblock In \emph{Proceedings of the 37th International Conference on Machine
  Learning}, volume 119 of \emph{Proceedings of Machine Learning Research},
  pages 6905--6916. PMLR.

\bibitem[{Nguyen et~al.(2022)Nguyen, Tran, Nguyen, Van~Huynh, Luu, and
  Nguyen}]{https://doi.org/10.48550/arxiv.2203.11400}
Kiet~Van Nguyen, Son~Quoc Tran, Luan~Thanh Nguyen, Tin Van~Huynh, Son~T. Luu,
  and Ngan Luu-Thuy Nguyen. 2022.
\newblock \href {https://doi.org/10.48550/ARXIV.2203.11400} {{VLSP} 2021 -
  {V}i{MRC} challenge: Vietnamese machine reading comprehension}.

\bibitem[{Raffel et~al.(2020)Raffel, Shazeer, Roberts, Lee, Narang, Matena,
  Zhou, Li, and Liu}]{JMLR:v21:20-074}
Colin Raffel, Noam Shazeer, Adam Roberts, Katherine Lee, Sharan Narang, Michael
  Matena, Yanqi Zhou, Wei Li, and Peter~J. Liu. 2020.
\newblock \href {http://jmlr.org/papers/v21/20-074.html} {Exploring the limits
  of transfer learning with a unified text-to-text transformer}.
\newblock \emph{Journal of Machine Learning Research}, 21(140):1--67.

\bibitem[{Rajpurkar et~al.(2018)Rajpurkar, Jia, and
  Liang}]{rajpurkar-etal-2018-know}
Pranav Rajpurkar, Robin Jia, and Percy Liang. 2018.
\newblock \href {https://doi.org/10.18653/v1/P18-2124} {Know what you don{'}t
  know: Unanswerable questions for {SQ}u{AD}}.
\newblock In \emph{Proceedings of the 56th Annual Meeting of the Association
  for Computational Linguistics (Volume 2: Short Papers)}, pages 784--789,
  Melbourne, Australia. Association for Computational Linguistics.

\bibitem[{Rajpurkar et~al.(2016)Rajpurkar, Zhang, Lopyrev, and
  Liang}]{rajpurkar-etal-2016-squad}
Pranav Rajpurkar, Jian Zhang, Konstantin Lopyrev, and Percy Liang. 2016.
\newblock \href {https://doi.org/10.18653/v1/D16-1264} {{SQ}u{AD}: 100,000+
  questions for machine comprehension of text}.
\newblock In \emph{Proceedings of the 2016 Conference on Empirical Methods in
  Natural Language Processing}, pages 2383--2392, Austin, Texas. Association
  for Computational Linguistics.

\bibitem[{Seo et~al.(2016)Seo, Kembhavi, Farhadi, and
  Hajishirzi}]{DBLP:journals/corr/SeoKFH16}
Min~Joon Seo, Aniruddha Kembhavi, Ali Farhadi, and Hannaneh Hajishirzi. 2016.
\newblock \href {http://arxiv.org/abs/1611.01603} {Bidirectional attention flow
  for machine comprehension}.
\newblock \emph{CoRR}, abs/1611.01603.

\bibitem[{Shinoda et~al.(2023)Shinoda, Sugawara, and
  Aizawa}]{shinoda2022shortcut}
Kazutoshi Shinoda, Saku Sugawara, and Akiko Aizawa. 2023.
\newblock \href {https://doi.org/10.1609/aaai.v37i11.26590} {Which shortcut
  solution do question answering models prefer to learn?}
\newblock \emph{Proceedings of the AAAI Conference on Artificial Intelligence},
  37(11):13564--13572.

\bibitem[{Sugawara et~al.(2018)Sugawara, Inui, Sekine, and
  Aizawa}]{sugawara-etal-2018-makes}
Saku Sugawara, Kentaro Inui, Satoshi Sekine, and Akiko Aizawa. 2018.
\newblock \href {https://doi.org/10.18653/v1/D18-1453} {What makes reading
  comprehension questions easier?}
\newblock In \emph{Proceedings of the 2018 Conference on Empirical Methods in
  Natural Language Processing}, pages 4208--4219, Brussels, Belgium.
  Association for Computational Linguistics.

\bibitem[{Sulem et~al.(2021)Sulem, Hay, and Roth}]{sulem-etal-2021-know-dont}
Elior Sulem, Jamaal Hay, and Dan Roth. 2021.
\newblock \href {https://doi.org/10.18653/v1/2021.findings-emnlp.385} {Do we
  know what we don{'}t know? studying unanswerable questions beyond {SQ}u{AD}
  2.0}.
\newblock In \emph{Findings of the Association for Computational Linguistics:
  EMNLP 2021}, pages 4543--4548, Punta Cana, Dominican Republic. Association
  for Computational Linguistics.

\bibitem[{Tran et~al.(2023)Tran, Do, Le, and
  Kretchmar}]{tran-etal-2023-impacts}
Son~Quoc Tran, Phong Nguyen-Thuan Do, Uyen Le, and Matt Kretchmar. 2023.
\newblock \href {https://aclanthology.org/2023.eacl-main.113} {The impacts of
  unanswerable questions on the robustness of machine reading comprehension
  models}.
\newblock In \emph{Proceedings of the 17th Conference of the European Chapter
  of the Association for Computational Linguistics}, pages 1543--1557,
  Dubrovnik, Croatia. Association for Computational Linguistics.

\bibitem[{Vaswani et~al.(2017)Vaswani, Shazeer, Parmar, Uszkoreit, Jones,
  Gomez, Kaiser, and Polosukhin}]{10.5555/3295222.3295349}
Ashish Vaswani, Noam Shazeer, Niki Parmar, Jakob Uszkoreit, Llion Jones,
  Aidan~N. Gomez, \L{}ukasz Kaiser, and Illia Polosukhin. 2017.
\newblock Attention is all you need.
\newblock In \emph{Proceedings of the 31st International Conference on Neural
  Information Processing Systems}, NIPS'17, page 6000–6010, Red Hook, NY,
  USA. Curran Associates Inc.

\end{thebibliography}
\bibliographystyle{acl_natbib}

\end{document}